# Action Selection Properties
# in a Software Simulated Agent

Carlos Gershenson García[1]
Instituto de Química/UNAM
Fundación Arturo Rosenblueth
Facultad de Filosofía y Letras/UNAM
carlos@jlagunez.iquimica.unam.mx

Pedro Pablo González Pérez
Instituto de Investigaciones Biomédicas/UNAM
Instituto de Química/UNAM
ppgp@servidor.unam.mx

José Negrete Martínez
Instituto de Investigaciones Biomédicas/UNAM
Maestría en Inteligencia Artificial/UV
jnegrete@mia.uv.mx

*Abstract*

*This article analyses the properties of the Internal Behaviour network, an action selection mechanism previously proposed by the authors, with the aid of a simulation developed for such ends. A brief review of the Internal Behaviour network is followed by the explanation of the implementation of the simulation. Then, experiments are presented and discussed analysing the properties of the action selection in the proposed model.*

## 1.Introduction

By the middle 1980's, researchers in the areas of artificial intelligence (AI), computer sciences, cognitive sciences and psychology realized that the idea of computers as intelligent machines was inappropriate. Inappropriate because the brain functions, far from being related with mathematical proofs and programmes execution, are related with the control of behaviour. Most researchers now agree that intelligence is manifested in behaviour (Pfeifer and Scheier, 1999). This has given place a research area to develop known as Behaviour-based Systems (BBS) (Brooks, 1986), in order to model intelligence in a different approach than previous Knowledge-based Systems (KBS).
This new line of research, also known as "autonomous agents", was inspired mainly in ethology, the branch of biology that studies animal behaviour. The term "agent" has been used in many other areas, but in this paper, when we talk about autonomous agents, we refer to the ones proposed by BBS. In BBS, the interaction with the problem domain is direct, while in KBS it is more limited. An autonomous agent perceives its problem domain through its sensors and actuates over it through its actuators. The problem domain of an autonomous agent is commonly a dynamic, complex and unpredictable environment, in which the agent tries to satisfy a set of goals or motivations, which can vary in time. An autonomous agent decides by himself how to relate his external and internal inputs with his motor actions in such a way that its goals may be achieved (Maes, 1994).
To satisfy his goals, an autonomous agent must select, at every moment in time, the most appropriate action among all possible actions that he could execute. This is what, in the context of BBS, is known as the action selection problem (ASP). While the ASP refers to what action, the agent (robot, animat or artificial creature) must select every moment in time; an action selection mechanism (ASM) specifies how these actions are selected. An ASM is a computational mechanism that must produce as an output a selected action when different external and/or internal stimuli have been given as inputs. In this point of view, an ASP indicates *which*, when an ASM indicates *how*.
Among the principal ASM and related works with the action selection are the hierarchical network of centres of Tinbergen (Tinbergen, 1950; Tinbergen, 1951), the psycho-hydraulic model of Lorenz (Lorenz, 1950; Lorenz, 1981), the hierarchical nodes network of Baerends (Baerends, 1976), the subsumption architecture of Brooks (Brooks, 1986;

---
[1] To whom all correspondence should be sent: Laboratorio de Qu mica Te rica. Instituto de Qu mica. Circuito Universitario, Coyoac n, M xico C.P. 04510 tel. (52) 5 622 4424

Brooks, 1989), the connectionist model of Rosenblatt and Payton (Rosenblatt and Payton, 1989), the bottom-up mechanism of Maes (Maes, 1990; Maes, 1991), the neuronal model of Beer (Beer, 1990; Beer, Chiel and Sterling, 1990), the neuroconnector network of Halperin (Hallam, Halperin and Hallam, 1994), the neuro-humoral model of Negrete (Negrete and Martínez, 1996), and the recurrent behaviour network of Goetz (Goetz and Walters, 1997). A complete revision of these ASMs is presented in (González, 1999), along with a comparison with the proposed ASM. These mechanisms have been inspired in models belonging to disciplines such as ethology, psychology, cognitive sciences, robotics, engineering, artificial neural networks and AI, among others. Some of these mechanisms contemplate completely the ASP, while some of them only deal with part of the problem.

In this work, the action selection properties of a proposed model by the authors (González, 1999) are illustrated in a simulation of a robot created for such effect. The proposed model was based in a distributed blackboard architecture, which, given its great capacity for coordination and integration of many tasks in real time and its extreme flexibility for the incorporation of new functionality, eases the implementation of the model and the incremental incorporation of new properties and learning processes, which enrich the action selection, making it more adaptive.

Some of the properties observed in the developed ASM are: (1) the strong dependence of the observed external behaviour in the internal states of the entity, (2) the action selection to the extern medium and to the internal medium, (3) the stability in the action selection, (4) the persistence in the execution of an action, (5) the existence of an external behaviour oriented to the search of a specific signal, and (6) the explicit relationship between the action selection and the learning processes.

This article is structured as follows: in the next section the main structural and functional characteristics of the ASM that we have proposed are presented: the internal behaviour network built with blackboard nodes (González, 1999). In section 3, the developed simulation is described: the animat, the simulated environment, and the behaviours that can be executed by the entity. Finally, section 4 presents and discusses some of the experiments developed in order to verify when the internal behaviour network was able to produce the effects claimed by it.

## 2. The Internal Behaviour Network

We have named "internal behaviour network" (IBeNet) to the action selection mechanism we have developed. A blackboard node is a blackboard system (Nii, 1989) with elements defined in the REDSIEX (González and Negrete, 1997) and ECN-MAES (Negrete and González, 1998) architectures. Although the blackboard architecture was developed in the area of KBS, its properties suit BBS as well. A complete explanation of the functionality of the IBeNet can be found in (González, 1999). The IBeNet architecture can be seen in Figure 1.

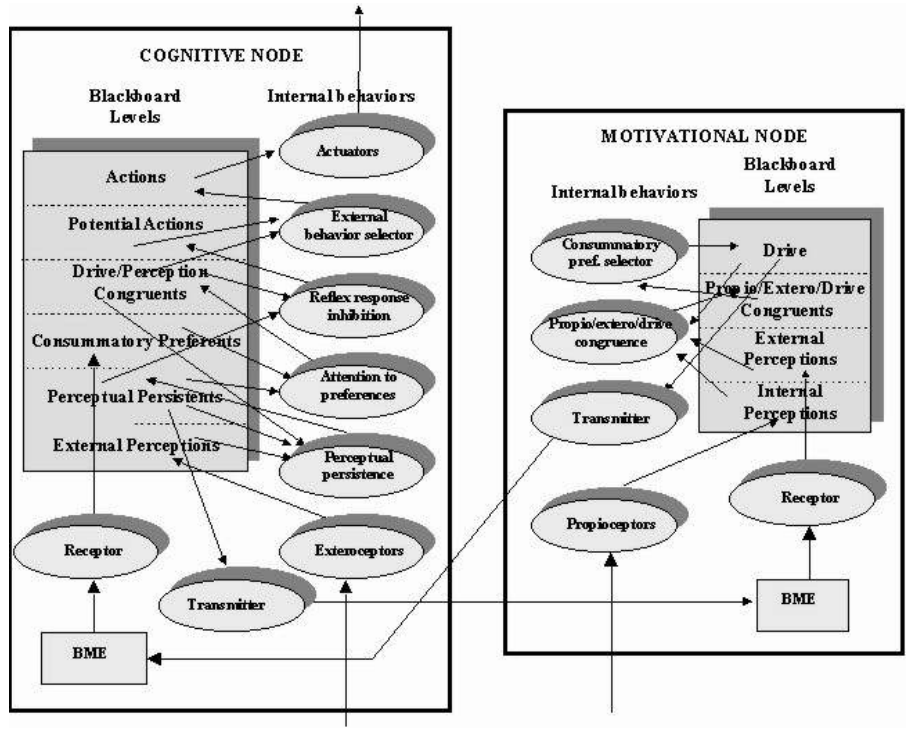

**Figure 1. Structure of the blackboard node network. Here have been omitted the activity state registers.**

As shown in Figure 1, the actual structure of the IBeNet defines two blackboard nodes: the cognitive node and the motivational node. The internal tasks required for the control of the action selection can only be satisfied over the base of the cooperative work between both nodes. The inputs of the exteroceptors come from the perceptual system, while the inputs of the propioceptors come from the internal medium. The outputs of the actuators are directed to the motor system.

### 2.1. Elements of a Blackboard Node

The structure of a blackboard node is defined by five basic elements: the blackboard, the internal behaviours, the activity state registers (REACs) of the internal behaviours, the interface/communication mechanisms, and the competition mechanism.

The blackboard is a shared data structure over which the internal behaviours execute their final actions, and by which the internal behaviours communicate. The interface/communication mechanisms also operate over the blackboard. The internal behaviours produce changes in the blackboard, which incrementally lead to the formation of a solution of the ASP.

The term "internal behaviour" is used to describe the intrinsic actions of the information process that occur at a blackboard node level. This is, inside the entity. In this sense, an internal behaviour may be of two kinds: the ones that generate actions directed to the extern medium of the entity, and those directed to the internal medium of the entity. From the structural point of view, an internal behaviour is defined as a package of simpler mechanisms called elemental behaviour, commonly from the same kind, and structured as production rules (if <condition> then <action>).

When the conditions of an elemental behaviour are satisfied, this is activated creating an activity state register (REAC). A REAC is a data structure that describes the characteristics of the action activated by the behaviour. When a REAC is selected by the competition mechanism, the action of the internal behaviour associated to the REAC is executed, and a new solution element will be created over the blackboard, or the certainty value of an existent solution element will be modified.

Over the blackboard the interface/communication mechanisms also operate. Three kinds of interface mechanisms have been defined: (1) the exteroceptors, which establish the interface between the perceptual system and the cognitive node; (2) the propioceptors, which establish the interface between the internal medium and the motivational node; and (3) the actuators, which define the interface between the cognitive node and the motor system. The communication mechanisms are the ones in charge of executing the reception and transmission of signals from and to other nodes in the network. (In ECN-MAES (Negrete and González, 1998) this are the behaviours with social knowledge).

A list of REACs (L-REAC) is the set defined by all the REACs associated to elemental behaviours of the same kind. To the level of each L-REAC a competitive process is established between the REACs in order to decide which elemental behaviour(s) will be winner(s). Only winner behaviours will be able to execute their final action over the blackboard. The competitive process can be of two kinds: at the end of the competition more than one behaviour may win, or only one behaviour may be proclaimed winner (the winner takes all).

### 2.2. The Cognitive Node

For the cognitive node, the following internal behaviours have been defined: perceptual persistence, attention to preferences, reflex response inhibition, and external behaviours selector; the interface mechanisms exteroceptors and actuators; and the communication mechanisms receptor and transmitter.

The cognitive node receives signals from the perceptual system through the exteroceptors and of the motivational node through the receptor mechanism; and sends signals to the motivational node through the transmitter mechanism and to the motor system through the actuators. The cognitive node role contains the processes of representation of perceptual signals, integration of internal and external signals, reflex response inhibition, and selection of the external behaviour that best fits to the actual internal state.

As can be seen in Figure 1, the domain blackboard of the cognitive node organizes the solution elements in six abstraction levels: external perceptions, perceptual persistents, consummatory preferents, drive/perception congruents, potential actions, and actions.

### 2.3. The Motivational Node

The motivational node receives signals from the internal medium through the propioceptors and from the cognitive node through the receptor mechanism; and sends signals to the cognitive node through the transmitter mechanism. It has defined the following internal behaviours: propio/extero/drive congruence and consummatory preferences selector.

The role of the motivational node contains the combination of internal and external signals, and the competition to the motivational level of motivationally incompatible behaviours. This is, the final observed external behaviour in the entity is strongly dependent of its internal states. All the internal states, for which there are useful external signals compete between them to determine the final external behaviour that the entity will execute. The competition is of the winner takes all type.

The domain blackboard of the motivational node organizes the solution elements in four abstraction levels: internal perceptions, external perceptions, propio/extero/drive congruents, and drive.

## 3. The Simulation

In order to verify when the IBeNet was able to produce by itself the effects claimed by it, a computer programme was written implementing this model, being used by an autonomous mobile robot (animat) simulated in virtual reality. The simulated robot was inspired in an original idea conceived by Neg rete (Negrete and Martínez, 1996).

This virtual reality simulation can be accessed via Internet at the following web address: http://132.248.11.4/~carlos/asia/animat/animat.html

### 3.1 The Simulated Robot

As can be seen in Figure 2, the graphical representation of the animat is a cone with a semicircumscript sphere. The tip of the cone indicates the animat's direction.

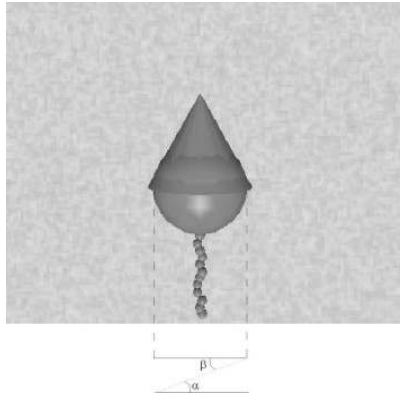

**Figure 2. Graphical representation of the animat**

The internal structure of the animat can be described in terms of four basic components: the perceptual system, the internal medium, the action selection mechanism, and the motor system. In Figure 3 this components and its interrelations can be appreciated.

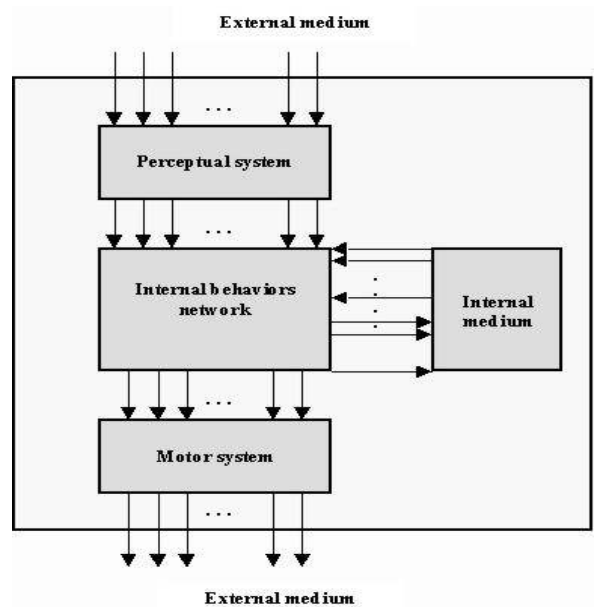

**Figure 3. Components that integrate the internal structure of the animat and its interrelations.**

### 3.1.1. The Perceptual System

The animat is situated in a plane (z,x) of a space (x,y,z). In this plane there can be found various kinds of external stimuli: water, represented by blue circles; food, represented by green spheres; grass, represented by texturized green circles; obstacles, represented by brown parallelepipeds; blobs (aversive stimuli), represented by black ellipsoids; and red and yellow spots (initial neutral stimuli), represented by circles of the respective colour. The animat only perceives those external stimuli that are inside the perceptual region ($R_p$) defined by the semicircle determined by (1).

$$R_p = \begin{cases} ((z-z_a)^2+(x-x_a)^2<r_p^2) \cap (x>x_a+\tan((th+\pi/2)*(z-z_a))) & \text{if } 0<th\leq\pi \\ ((z-z_a)^2+(x-x_a)^2<r_p^2) \cap (x<x_a+\tan((th+\pi/2)*(z-z_a))) & \text{if } \pi<th\leq2\pi \end{cases} \quad (1)$$

The expression $(z-z_a)^2+(x-x_a)^2<r_p^2$ determines the set of all points contained in the circle with centre in $(z_a,x_a)$ and radius $r_p$ ($r_p$ stands for perception radius). The expression $x>x_a+\tan((th+\pi/2)*(z-z_a))$ determines the set of all points above the straight line perpendicular to the animat's orientation (th) and which crosses the point $(z_a,x_a)$, while the expression $x<x_a+\tan((th+\pi/2)*(z-z_a))$ determines the set of all points below of such straight line. When $0 < th \leq \pi$, the animat's perceptual region $R_p$ is the semicircle defined by the intersection $((z-z_a)^2+(x-x_a)^2<r_p^2)\cap(x>x_a+\tan((th+\pi/2)*(z-z_a)))$; while when $\pi < th \leq 2\pi$, $R_p$ is defined by the intersection $((z-z_a)^2+(x-x_a)^2<r_p^2)\cap(x<x_a+\tan((th+\pi/2)*(z-z_a)))$. Once determined the stimuli, which fall inside the animat's perceptual region, those stimuli that are behind an obstacle, are eliminated. In addition, the radius of perception $r_p$ is proportional to the lucidity of the animat. The animat's perceptual region can be appreciated in Figure 4.

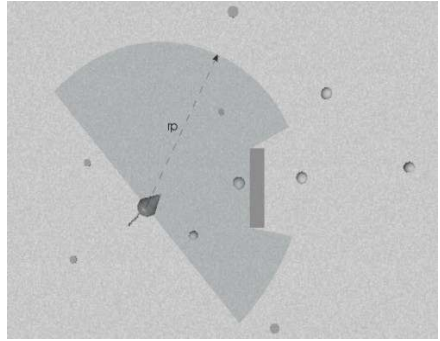

**Figure 4. Animat's perceptual region.**

The perceptual system captures the magnitude of the stimuli and the distance between them and the animat. Then, the perceptual system sends to the ASM a pondered value corresponding to every kind of external stimuli, which is a function of the ratio between magnitude and distance for all the stimuli of the same kind. The pondered value represents the strength with which the kind of stimuli will be registered by the exteroceptors in the external perceptions blackboard level of the cognitive node.

Once the animat stops perceiving a stimulus (it is left behind the animat or an obstacle), the stimulus reverberates shortly in the perceptual system, decreasing its value, until it is very small and the system "forgets" the stimulus. This simulates, in the perceptual system, a short-medium time memory.

### 3.1.2. The Motor System

As can be seen in Figure 2, the walk of the animat is commanded by the angular steps $\alpha$ and $\beta$ with centre in the extremes of the diameter of the projection of the animat's sphere in the plane (z,x) perpendicular to the animat's orientation. The wander external behaviour is produced when angles $\alpha$ and $\beta$ take random values between zero and one (radians), which describes a random trajectory. On the other hand, the exploration oriented to the search of a specific signal external behaviour is produced when angles $\alpha$ and $\beta$ take equal values, which describes the animat's trajectory as a straight line. The size of the steps is proportional to the animat's strength.

### 3.1.3. The internal medium

The animat's internal medium is expressed through a set of variables, in which each one of them represents an internal state or need. For the actual simulation, the internal states strength, lucidity, security, fatigue, thirst and hunger have been considered. As it can be seen in Figure 3, the IBeNet is the only component of the animat's internal structure that establishes a direct interface with the internal medium. The IBeNet perceives each instant the actual value of each of the variables of the internal medium through the propioceptors; and updates such value in a determined proportion when the external behaviour associated to the internal state has been executed. For example, the execution of the consummatory behaviours eat and drink will reduce the values of the internal states hunger and thirst, respectively.

When the internal states hunger, thirst and/or fatigue are very high, the internal states strength and lucidity begin to decrease, slowing the movements of the animat and affecting his perception. On the other hand, if the internal states hunger, thirst and/or fatigue are satisfied, then strength and lucidity restore slowly. When strength reaches a value of zero, the animat dies.

## 3.2. The Simulated Environment

The animat's environment is defined by a plane (z,x) of a space (x,y,z). The plane (z,x) is delimited by a frame. In the area defined by this frame different objects can be created. This objects represent the external stimuli food (green spheres), water (blue circles), grass (texturized green circles), fixed obstacles (brown parallelepipeds), blobs (black ellipsoids), and other kinds of stimuli that initially have no specific meaning for the entity (red and yellow circles). The frame that delimits the plane (z,x) is also considered as a fixed obstacle. Figure 5 shows an air view of the simulated environment.

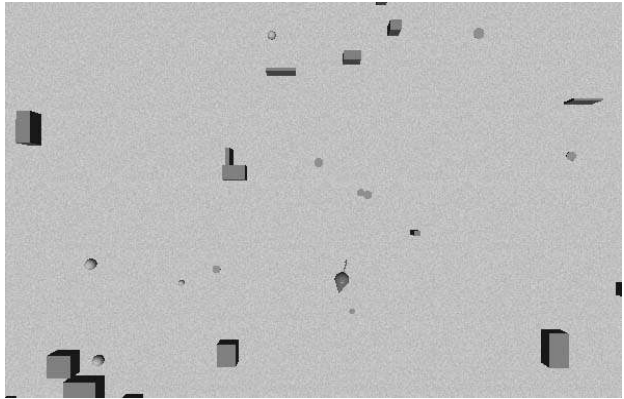

**Figure 5. Air view of the simulated environment**

## 3.3. The Behaviour Repertory

The behaviour repertory (external actions) that the animat can execute is shown in Table 1. The selection of these behaviours responds to the properties of the IBeNet that are proved. Many of these behaviours can be executed only when both conditions, internal and external, have been satisfied. This is, there is a high internal need and the external input capable of satisfying that need has been perceived. This is the case of the behaviours approach food, eat, approach water, drink, approach food and water, approach grass, sleep and runaway.

| Behaviour | External input | Internal input |
|---|---|---|
| Avoid obstacle | Obstacle at range | None |
| Wander | None | None |
| Explore | None | Thirst and/or hunger |
| Approach food | Food perceived | Hunger |
| Eat | Food at range | Hunger |
| Approach water | Water perceived | Thirst |
| Drink | Water at range | Thirst |
| Approach food and water | Food and water perceived | Hunger and thirst |
| Approach grass | Grass perceived | Fatigue |
| Rest | Grass at range | Fatigue |
| Runaway | Blob perceived | Safety |

**Table 1. Animat's behaviours repertory**

According to McFarland (McFarland and Houston, 1981; Maes, 1991), the associated behaviours with one or more motivations (or internal needs) are known as consummatory behaviours, while the ones that are not associated directly with some motivation are known as appetitive behaviours. In this way, a consummatory behaviour is that which an animat (or artificial entity) really wants to satisfy when the motivation associated to this is high; while an appetitive behaviour only contributes so that the consummatory behaviour can be executed.

In the animat simulation, we have considered that both kinds of behaviour exist: consummatory and appetitive. However, in difference to McFarland (McFarland and Houston, 1981), we have assumed that consummatory and appetitive behaviours are directly associated with some motivation. The difference between both kinds of behaviours lies in that a consummatory behaviour is that which executes the final action required by the animat to satisfy a high internal need, and commonly reduces the strength of such need; while an appetitive behaviour contributes so that the consummatory behaviour can be executed, but the execution of the first does not reduce directly the motivation level to which it is associated. For example, if we consider that the approach water behaviour is appetitive and the drink behaviour is consummatory, then both behaviours will be associated to the thirst internal need; and the animat will only approach the water if it is thirsty, contributing in this way so that the desired final action, drink, will be executed.

## 4. Experiments

Let us consider an environment as the one described in section 3.2, in which there have been created objects $O_i^+$ that have a specific meaning for the animat, such as water, food, fixed obstacles, etc.; and objects $O_i^n$, which can acquire a new meaning for the animat. Let the positions in the environment, described as a plane, be defined by the pair $(z, x)$. Let the entity animat defined by (1) a set of primitive actions that this can execute, as the ones described in section 3.1.2; (2) an ASM, as the internal behaviour network described in section 2; (3) a perceptual system, as the one described in section 3.1.1; and (4) a set of sensors of needs, that perceive the states of the internal medium.

The initial experiments were developed in order to verify: (1) the influence of the internal states in the observed external behaviour of the animat, (2) the role of the competition at a motivational level in the selection of the external behaviour to execute, (3) the exploratory behaviour oriented to the search of a specific signal and the reduction of the response times of the animat, (4) the stability in the selection and persistence in the execution of the external behaviours, (5) the discrimination between different stimuli taking in count the quality of them, (6) avoid aversive stimuli, (7) the non persistence in the execution of a consummatory action when an aversive stimulus is perceived, and (8) the role of the learning processes in the action selection. Next, experiments (2), (5), and (7) are presented and discussed.

### 4.1. The Role of the Competition at a Motivational Level in the Selection of the External Behaviour to Execute

Among all consummatory preferences selector elemental behaviours that have satisfied their condition, a competition of the winner takes all type takes place, in order to decide which of these behaviours will determine the external action that the animat will later execute. The behaviours that participate in this competition are incompatible behaviours, in the sense that only one will be able to send a drive signal directed to the cognitive node. A complete explanation of this competition is given in (González, 1999).

In order to illustrate this, a situation has been modelled, in which the animat has different internal states with significant and different values; at the same time that it is perceiving external signals capable to satisfy each one of these internal states. This modelled situation can be seen in Figures 6 and 7.

Figure 6 shows an initial state given by the position of the animat in the environment, the position and magnitude of external stimuli, and the initial values of the internal states thirst, hunger and fatigue. This is, the initial state defines the existent situation before the competition process at a motivational level takes place. Note that there is a need of drinking due to the very high value of thirst. As shown in Figure 6, the animat is perceiving plenty sources of food, water, and places where to rest.

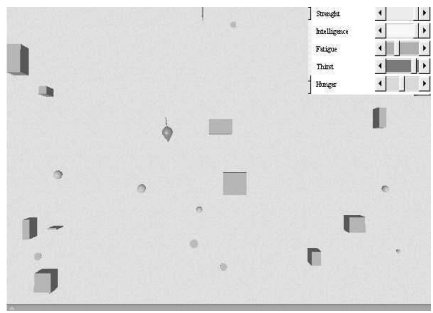 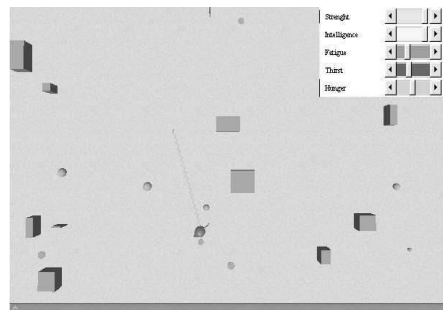

**Figure 6. Animat's initial state. There are thirst, hunger and fatigue; and there are sources of water, food and resting places.**

**Figure 7. State reached once the external behaviours approach water and drink were executed.**

Since all of the animat's actual internal needs can be satisfied, because the corresponding external stimuli have been perceived, a competitive process takes place between the consummatory preferences selector elemental behaviours, in order to decide which behaviour must be executed. In Figure 7 a state is shown, in which the approach water appetitive behaviour was executed until the water source was at range, and the animat could execute the drink consummatory behaviour. In consequence of this, thirst was reduced.

Therefore, as it has been illustrated in Figures 6 and 7, every time that two or more external behaviours satisfy their conditions, a competition at a motivational level will take place in order to decide which is the most appropriate external behaviour for the animat to execute.

### 4.2. Discrimination Between Different Stimuli Taking in Count the Quality of Them

Let us have an initial state as the one shown in Figure 8, in which the animat has thirst and hunger, as much as thirst as hunger, and the animat perceives sources of water and food. In particular, it perceives an isolated source of water, an isolated source of food, and a source of food next to a source of water. Since these sources of water and food are next to each other, we consider both sources as a new kind of stimulus: source of water and food.

The "discrimination between different stimuli taking in count the quality of them" property establishes for the previous situation, that the best of this three stimuli is the source of water and food for the actual states of thirst and hunger of the animat. Once the competition at a motivational level takes place, the first external behaviour that the animat executes is approach water and food.

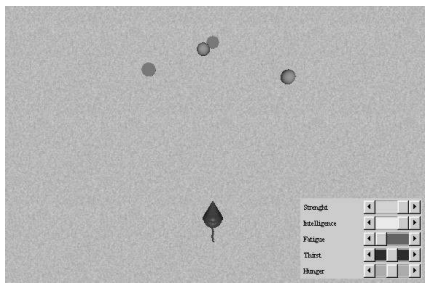

**Figure 8. Animat's initial state. There is thirst and hunger, as much thirst as hunger; and the animat perceives a source of food, a source of water, and a source of water and food.**

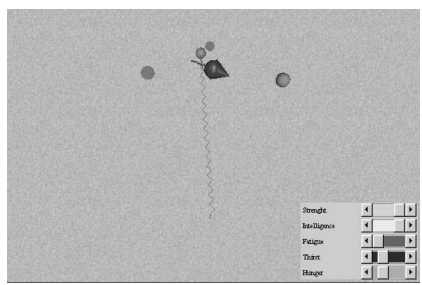

**Figure 9. State reached once the behaviours approach food and water, drink, and eat have been executed.**

In Figure 9 is shown a state where the approach food and water external behaviour was executed until the source of water and food was at range for the animat. Then, the drink behaviour was executed, since the consummatory preferences selector elementary behaviour associated to the thirst internal state won the competition at a motivational level. As a result of the persistence of the drink consummatory action, the strength of the thirst internal state was decreased, which allowed the consummatory preferences selector elemental behaviour associated to the hunger internal state to win the competition at a motivational level, and as a result, the eat external behaviour was executed, until the strength of the hunger internal state would decrease enough.

The direct consequence that the animat had considered the source of water and food stimulus as the one with the best quality among all the stimuli perceived at that moment, was a reduction in the amount of external actions that the animat would need to execute in order to satisfy both his internal needs thirst and hunger. This is, once the thirst internal state was satisfied, the animat did not have to execute the approach food external behaviour in order to execute the eat behaviour, and satisfy the hunger internal state; since the food was at range and the eat behaviour could be executed without the precedence of an appetitive external behaviour.

### 4.3. The Non persistence in the Execution of a Consummatory Action when an Aversive Stimulus is Perceived

The persistence in the execution of the external behaviours means that once the execution of the external behaviour has been initiated, the animat will always try to finish this, avoiding that other stimuli, irrelevant for the moment, distract his attention. Nevertheless, the non persistence in the execution of a consummatory action will take place when an aversive stimulus (blob) has been perceived and the ratio between the magnitude of the blob and its distance to the animat will constitute a risk proportional to the given degree of the safety internal state of the animat.

In order to exemplify this property, let us consider an initial state as the one shown in Figure 10, in which the animat has a high degree of thirst and it is perceiving a water source, but has not perceived an aversive stimulus that is also in the environment.

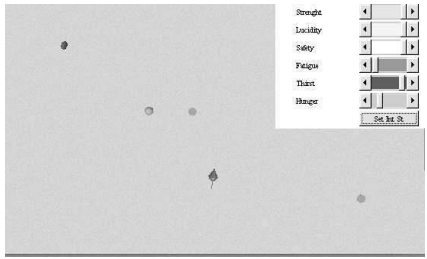 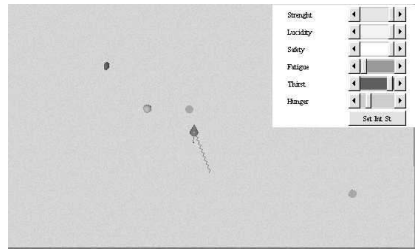

**Figure 10. Animat's initial state. There is a high thirst and water perceived.**

**Figure 11. The approaching water behaviour is being executed. The animat has not perceived the blob yet.**

In Figure 11, a state is shown where the approach water behaviour is being executed, so that the animat is closer to the water source, but it still has not perceived the blob, which has moved and it is also nearer the water source.

As it can be seen in Figure 12, the animat continued the execution of the approach water behaviour, and once that this was at range, it began to execute the drink behaviour, which had to be interrupted before it fully satisfied the thirst need, due to the proximity of the blob. Therefore, the external runaway behaviour was executed. In Figure 12, it can be seen that the internal state thirst was decreased because of the execution of the drink behaviour, but it was not persistent enough to satisfy the need completely.

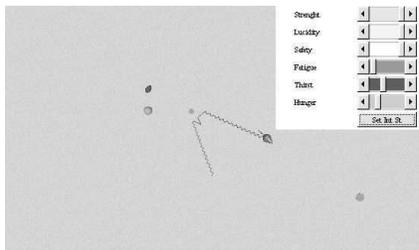 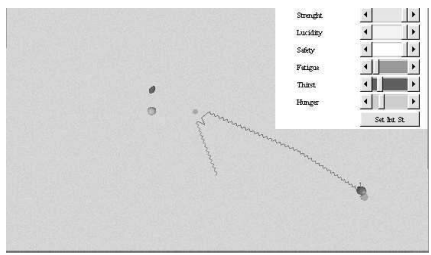

**Figure 12. The action drink is interrupted due to the proximity of the blob. The runaway action is executed.**

**Figure 13. Behaviours pattern that shows all the external actions executed by the animat.**

As it is shown in Figure 13, once far of the blob, the animat begins to explore again in search of a water source in order to complete the consummation of the drink behaviour. A water source is perceived, the animat approaches it, and completes the interrupted action due to the presence of the aversive stimulus.

## 5. Conclusions and Future Work

The ASM that implements the IBeNet is characterized by the following properties, some of which could be appreciated in the experiments presented here, and the rest were verified in previous work (González, 1999):

- The external behaviour is influenced by the entity's internal states.
- Motivationally incompatible behaviours compete between them and the competition is at a motivational level.
- Modulation of reactive behaviour.
- Goal-oriented behaviour.

- Non indecision in the action selection.
- Stability in the selection and persistence in the execution of the external behaviours.
- Regulated spontaneity.
- Satiation.
- Changes in responsiveness.
- Varying attention.
- Preactivation of internal behaviours.
- Discrimination between the different kinds of external stimuli taking in count the quality of them.
- The existence of an external default behaviour oriented to the search of a specific signal accomplishes smaller response times, in respect of the satisfaction of an imperious internal need.
- Associative learning (classical primary and secondary conditionings).
- Reflex Response Inhibition.

While the simulation explained here was able to verify the main properties that characterize the action selection in the IBeNet, the implementation of the mechanism constitutes itself an ideal scenario for the modelling and testing of new behaviours and desired properties in the action selection of an autonomous agent, proceeding from areas such as reactive robotics, ethology, and cognitive sciences.

An immediate application of the IBeNet will be in the biomedical sciences area, in the modelling, comprehension, and prediction of biological system. Specifically, in protein interactions.

Among the future works there is the modelling of an artificial society of agents, where each agent will implement the IBeNet in a similar way as the one proposed here; and the integration of a previously proposed model if emotions (Gershenson, 1999) to the IBeNet, which will enhance the action selection.

## 6. References

- Baerends, G. (1976) The functional organization of behaviour. *Animal Behaviour*, 24, 726-735.
- Beer, R. (1990) Intelligence as Adaptive Behaviour: an Experiment in Computational Neuroethology. Academic Press.
- Beer, R., Chiel, H. and Sterling, L. (1990) A biological perspective on autonomous agent design. *Robotics and Autonomous Systems*, 6, 169-186.
- Brooks, R. A. (1986) A robust layered control system for a mobile robot. *IEEE Journal of Robotics and Automation.* RA-2, April, pp. 14-23.
- Brooks, R. A. (1989) A robot that walks: Emergent behaviour from a carefully evolved network. *Neural Computation*, 1, 253-262.
- Gershenson, C. (1999) Modelling Emotions with Multidimensional Logic. *Proceedings of the 18th International Conference of the North American Fuzzy Information Processing Society* (NAFIPS '99), pp. 42-46. New York City, NY.
- Goetz, P. and D. Walters (1997) The dynamics of recurrent behaviour networks. *Adaptive Behaviour*, 6(2), 245-282.
- González, P.P. (1999) Redes de Conductas Internas como Nodos-Pizarrón: Selección de Acciones y Aprendizaje en un Robot Reactivo. *Tesis Doctoral*, Instituto de Investigaciones Biomédicas/UNAM, México.
- González, P.P. and J. Negrete (1997) REDSIEX: A cooperative network of expert systems with blackboard architectures. *Expert Systems*, 14(4), 180-189.
- Hallam, B.E, J. Halperin and J. Hallam (1994) An Ethological Model for Implementation in Mobile Robots. *Adaptive Behaviour*, 3 (1), pp 51-79.
- Lorenz, K. (1950) The comparative method in studying innate behaviour patterns. Symposia of the Society for Experimental Biology, 4, 221-268.
- Lorenz, K. (1981) Foundations of Ethology. Springer-Verlag.
- Maes, P. (1990) Situated agents can have goals. *Journal of Robotics and Autonomous Systems*, 6(1&2).
- Maes, P. (1991) A bottom-up mechanism for behaviour selection in an artificial creature. In J.A. Meyer and S.W. Wilson (ed.), *From Animals to Animats: Proceedings of the First International Conference on Simulation of Adaptive Behaviour* MIT Press/Bradford Books.
- Maes, P. (1994) Modelling Adaptive Autonomous Agents. *Journal of Artificial Life*, 1 (1,2), MIT Press.
- McFarland, D.J. and A.I. Houston (1981) Quantitative Ethology: The state space approach. Boston: Pitman.
- Negrete, J. and M. Martínez (1996) Robotic Simulation in Ethology. *Proceedings of the IASTED International Conference: Robotics and Manufacturing*, Honolulu, Hawaii, USA, pp. 271-274.
- Negrete, J. and P.P. González (1998) Net of multi-agent expert systems with emergent control. *Expert Systems with Applications*, 14(1) 109-116.

- Nii, H. P. (1989) Blackboard systems. En A. Barr, P. R. Cohen, y E. A. Feigenbaum (ed.), The Handbook of Artificial Intelligence, volume IV, Addison-Wesley Publishing Company
- Pfeifer, R. and C. Scheier (1999) Understanding Intelligence. MIT Press.
- Rosenblatt, K. and D. Payton (1989) A fine-grained alternative to the subsumption architecture for mobile robot control. *Proceedings of the IEEE/INNS International Joint Conference on Neural Networks,* IEEE.
- Tinbergen, N. (1950) The hierarchical organization of mechanisms underlying instinctive behaviour. *Experimental Biology*, 4, 305-312.
- Tinbergen, N. (1951) The Study of Instinct. Claredon Press.